\documentclass{article}

\usepackage[preprint]{neurips_2026}

\usepackage[utf8]{inputenc} 
\usepackage[T1]{fontenc}    
\usepackage{hyperref}       
\usepackage{url}            
\usepackage{booktabs}       
\usepackage{makecell}       
\usepackage{amsmath}        
\usepackage{amsfonts}       
\usepackage{amssymb}        
\usepackage{nicefrac}       
\usepackage{microtype}      
\usepackage{xcolor}         
\usepackage{graphicx}       
\usepackage{import}         

\usepackage{hyperref}
\title{Precision As You Need: Stochastic Computing Is a Dense Adaptive Quantizer}

\author{%
  \begin{minipage}[t]{\dimexpr\textwidth-2\tabcolsep\relax}
    \normalfont\raggedright
    \textbf{Haoran Jin}\textsuperscript{*,$\dagger$},
    \textbf{Kangqi Zhang}\textsuperscript{*},
    \textbf{Jirong Yang},
    \textbf{Barry Lyu}, 
    \textbf{Qiuyi Ding},
    \textbf{Ruijie Gao}, \\
    \textbf{Nathan Bleier} \\[0.5ex]
    Department of Electrical Engineering and Computer Science, University of Michigan \\
    \texttt{\{allenjin, zhkangqi, yjrcs, barrylyu, dingqy, ruijieg, nbleier\}@umich.edu}
  \end{minipage}%
}

\begin{document}

\maketitle

\begingroup
  \renewcommand{\thefootnote}{\fnsymbol{footnote}}
  \footnotetext[1]{Equal contribution.}
  \footnotetext[2]{Corresponding author: Haoran Jin (\texttt{allenjin@umich.edu}).}
\endgroup

\begin{abstract}
Matrix multiplications dominate the inference cost of modern transformer-based
vision models, yet existing efficiency techniques such as post-training
quantization and mixed-precision inference are largely limited to the small set
of fixed-width formats---INT4, INT8, BF16, and FP16---supported by conventional
accelerators.
We revisit \emph{stochastic computing} (SC) as a way to lift this constraint:
viewed as a \emph{dense adaptive quantizer}, SC controls precision by
bit-stream length~$L$ rather than a fixed datapath, while each multiplication
reduces to a single AND/XNOR gate.
We build a GPU library that emulates SC matrix multiplication at scale, exposes stream lengths as first-class kernel arguments, and evaluates SC
end-to-end on image classification, object detection and instance segmentation,
class-conditional image generation, and visual world-model planning.
On top of this substrate, we develop a dynamic per-row mixed-precision policy
that assigns stream length per token or group at matched average budget,
requires no retraining, and uses the same SC hardware across schedules.
Across tasks, SC remains competitive with fixed-format INT quantization at
matched bit budgets, while per-row mixed precision helps maintain accuracy at
lower average stream lengths. These results provide software-level feasibility
evidence that SC can serve as a dense-precision substrate for fine-grained
mixed-precision inference on modern vision transformers.

\end{abstract}

\section{Introduction}
\label{sec:intro}
The computational footprint of modern neural networks has grown faster than the efficiency gains provided by conventional hardware scaling.
Across vision, language, generation, and world-modeling workloads~\citep{dosovitskiy2020image,brown2020gpt3,touvron2023llama,peebles2023dit,ha2018world,hafner2023dreamerv3}, matrix multiplications dominate inference cost, complicating deployment in latency- and energy-constrained settings, including edge devices, real-time control systems, and power-limited datacenters.

The dominant response has been low-bit arithmetic that trades efficiency at the cost of accuracy.
Post-training quantization (PTQ) maps weights and activations to INT8, INT4, or lower precision, exploiting the empirically observed robustness of trained networks to numerical perturbations~\citep{nagel2021ptq,dettmers2022llmint8,frantar2023optq,xiao2023smoothquant}.
Mixed-precision methods extend this idea by assigning different bit-widths to different operators, layers, or tensors, thereby recovering accuracy in cases where a uniform low-precision format is too restrictive~\citep{wang2019haq,dong2019hawq,rouhani2023mxformat,bitmod}.

Conventional mixed precision, however, is constrained by hardware: each supported precision typically requires a distinct datapath or mode, so deployed accelerators expose only a small set of formats, such as INT4, INT8, BF16, and FP16.
This sparse, hardware-defined precision ladder limits the granularity of precision, thereby constraining the effectiveness of mixed-precision inference.

Stochastic computing (SC) provides a fundamentally different precision substrate: precision is determined by execution time rather than by a fixed hardware format.
In SC, a real value is represented by the ratio of $1$s in a stochastic bit stream~\citep{gaines1969stochastic,alaghi2013survey}.
Multiplication of two such streams reduces to a single bitwise AND gate in the unipolar case or XNOR gate in the bipolar case.

SC has three properties well-suited to neural-network inference: (i) \textbf{Area and energy efficiency.} A multi-bit multiplier is replaced by a single logic gate, enabling substantial reductions in area and energy relative to integer or floating-point MACs~\citep{alaghi2013survey,sim2017scnn,usystolic,cambriconu,exploit,printedsc} (ii) \textbf{Dense, hardware-shared precision.} SC precision is governed by the bit-stream length $L$, which can be \emph{any} positive integer rather than only a conventional word size. The same hardware can compute at different precisions by terminating streams at different lengths, allowing a single accelerator to support a dense set of precision choices and fine-grained mixed-precision schedules. (iii)  \textbf{Inherent error tolerance.} A bit flip in an SC stream perturbs the decoded value by only $1/L$, whereas an error in a high-order bit of an integer or floating-point word can induce a much larger deviation~\citep{surveysc,errorSC}. This property is particularly relevant for unreliable computing environments, including space computing.

Despite these advantages, prior work on SC for deep learning has remained limited in scope.
Most studies evaluate small CNNs or simplified transformer models~\citep{ren2017sc,sim2017scnn,usystolic,cambriconu}.
To our knowledge, no prior work has performed an end-to-end evaluation of SC across modern transformer-based AI workloads or quantitatively demonstrated how its dense precision substrate can be used for fine-grained mixed-precision inference.

This work addresses that gap by treating SC as a \emph{dense adaptive quantizer}.
We show that this perspective enables competitive accuracy with substantially reduced compute across four families of modern models.
Our contributions are: (i) \textbf{A scalable GPU library for SC simulation.} We present, to our knowledge, the first GPU implementation that can effectively emulate SC matrix multiplication at scale. The library implements bit-stream generation, AND/XNOR-based multiplication, and counter-based decoding as fused Triton kernels, and exposes per-row and per-group bit-stream lengths for algorithmic exploration. (ii) \textbf{An end-to-end evaluation across vision-centric transformer workloads.}
We evaluate SC inference on standard vision tasks---classification, detection, and
instance segmentation---as well as diffusion-based image
generation and learned world-model rollouts. To our knowledge, this is the first
end-to-end study showing that an SC-based compute pipeline can preserve task-level
performance across various transformer-based vision workloads.
(iii)  \textbf{A fine-grained mixed-precision algorithm.} We introduce a runtime mixed-precision scheme that assigns a separate bit-stream length to each \emph{row} (token or group) in every matrix multiplication, guided by lightweight operator-specific saliency metrics.

The GPU kernel is available at
\url{https://github.com/CrucibleComputingGroup/scmp_kernels}.


\section{Related Works}
\label{sec:related}
\subsection{Stochastic Computing}

Stochastic computing (SC)~\citep{gaines1969stochastic} encodes real values as
the density of $1$s in probabilistic bit streams, reducing multi-bit
multiplication to a single AND/XNOR gate and realizing addition, division, and
common activations with small finite-state machines; classical applications
include LDPC decoding and image processing~\citep{alaghi2013survey}. For neural
networks~\citep{liu2020scnnsurvey}, SC-DCNN~\citep{ren2017sc} and the SC
multiplier of Sim and Lee~\citep{sim2017scnn} scale SC to medium-sized CNNs
while preserving accuracy, and recent hardware substrates such as
uSystolic~\citep{usystolic} and Cambricon-U~\citep{cambriconu} push SC closer to
modern accelerator scale.

\subsection{Post-training Quantization and Mixed Precision}

Post-training quantization (PTQ) compresses pretrained networks by mapping
weights and activations to low-bit formats without retraining~\citep{nagel2021ptq}.
For LLMs, OPTQ~\citep{frantar2023optq}, LLM.int8()~\citep{dettmers2022llmint8},
SmoothQuant~\citep{xiao2023smoothquant}, and AWQ~\citep{lin2024awq} enable
low-bit transformer inference by addressing weight quantization and activation
outliers. For vision and generative transformers, PTQ4ViT~\citep{yuan2022ptq4vit},
Q-Diffusion~\citep{li2023qdiffusion}, and Q-DiT~\citep{chen2025qdit} adapt PTQ to
ViTs, diffusion models, and Diffusion Transformers. Mixed-precision methods further assign different precisions to different
layers, operators, or tensors. HAQ~\citep{wang2019haq}, HAWQ-V3~\citep{yao2021hawqv3},
and BRECQ~\citep{li2021brecq} search for sensitivity-aware bit-width assignments,
while MX formats~\citep{rouhani2023mxformat}, MPQ-DM~\citep{feng2025mpqdm}, and
MixDiT~\citep{kim2025mixdit} specialize mixed precision for modern accelerators
and generative workloads. 

\section{Preliminaries and Problem Formulations}
\label{sec:preliminaries}

\subsection{Stochastic Computing Basics}
\label{sec:sc_basics}

\paragraph{Encoding.}
Stochastic computing (SC) represents a real value by the probability that each bit of a single-bit binary sequence is $1$~\citep{gaines1969stochastic,alaghi2013survey}.
A \emph{unipolar} stream encodes $v\in[0,1]$ by a length-$L$ sequence $Z=(z_1,\ldots,z_L)\in\{0,1\}^L$ in which each bit is $1$ with probability $v$.
A \emph{bipolar} stream encodes $v\in[-1,1]$ with $\Pr[z_\ell=1]=(v+1)/2$.
Since neural-network activations and weights are signed, we use bipolar encoding throughout.

\begin{figure}[ht]
\centering
\begin{minipage}[b]{0.32\linewidth}
\centering
\includegraphics[width=\linewidth]{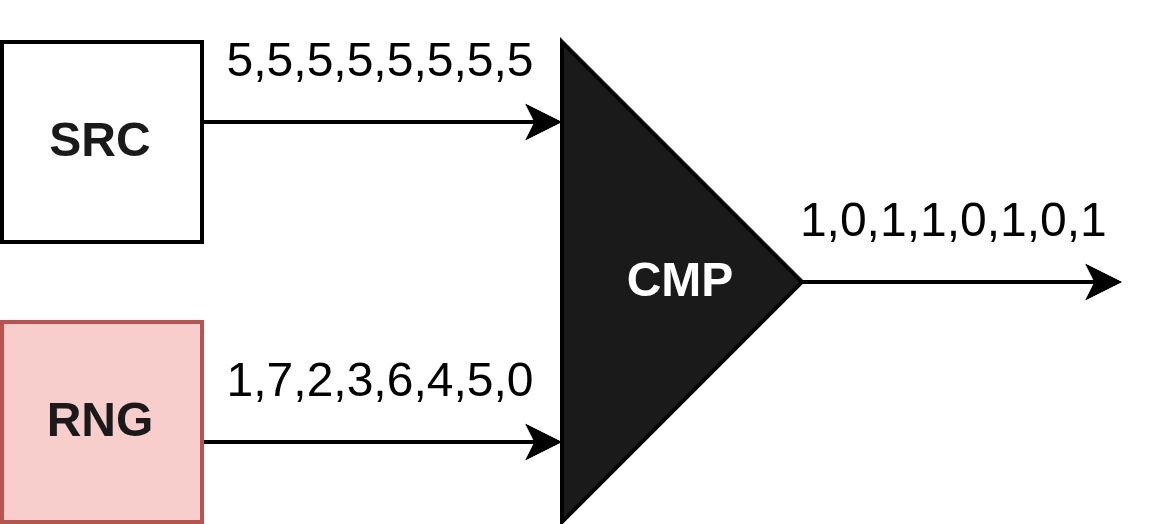}\\[2pt]
{\small (a) Bitstream generation}
\end{minipage}\hfill
\begin{minipage}[b]{0.32\linewidth}
\centering
\includegraphics[width=\linewidth]{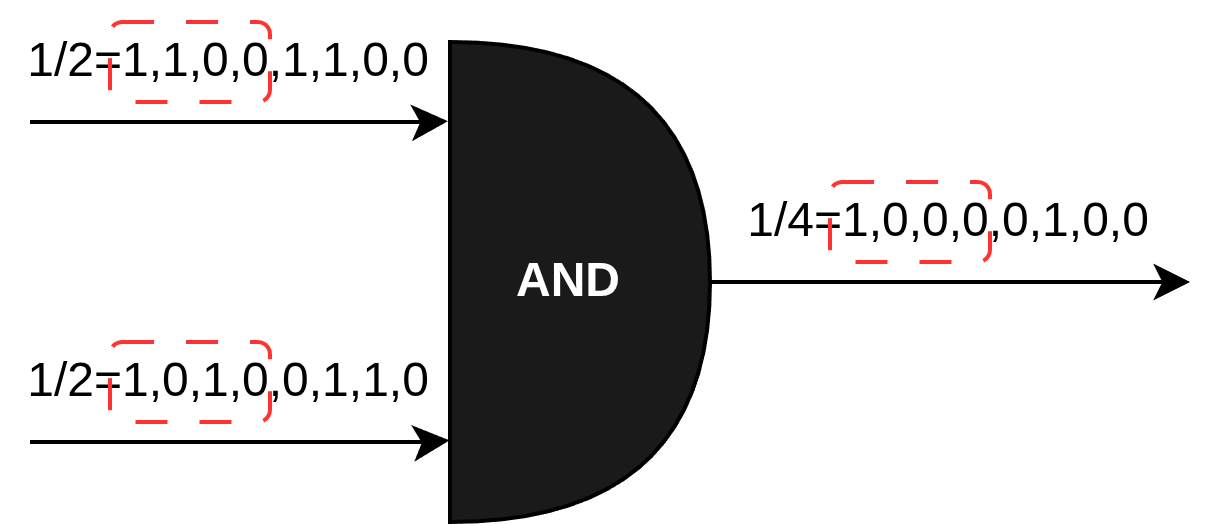}\\[2pt]
{\small (b) Bitstream multiplication}
\end{minipage}\hfill
\begin{minipage}[b]{0.32\linewidth}
\centering
\includegraphics[width=\linewidth]{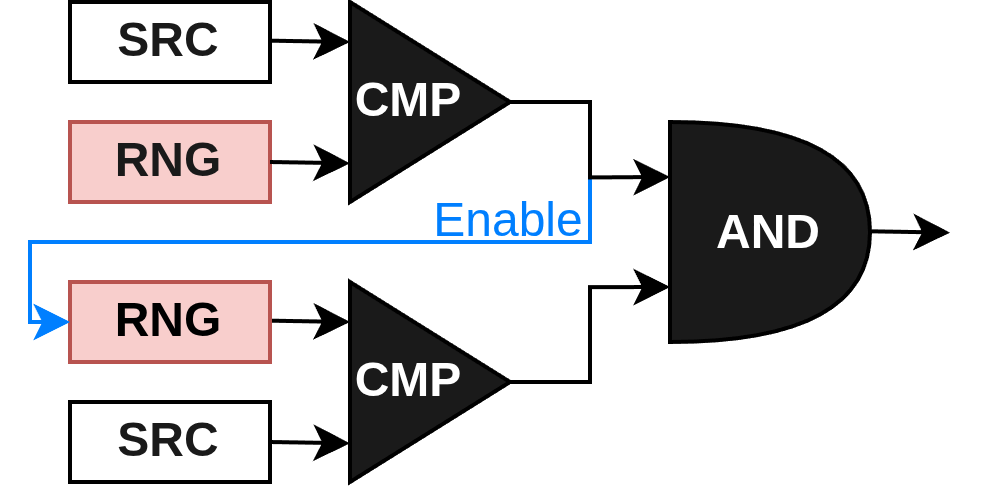}\\[2pt]
{\small (c) C-BSG multiplier}
\end{minipage}
\caption{Circuit primitives of stochastic computing.
\textbf{(a)} A SNG implemented as an integer comparator: at each cycle the source value is compared against an RNG draw, emitting a single bit.
\textbf{(b)} Stochastic multiplication via a bitwise gate: the unipolar product is realized by an AND gate whose output bit-probability factors as the product of the input marginals.
The red circle denotes early termination.
\textbf{(c)} Conditional bitstream generation (C-BSG): the second RNG is gated by the first stream so the conditional distribution of $z^B$ matches its marginal, enforcing $\mathrm{SCC}=0$ and making the bipolar XNOR multiplier exact at finite $L$.}
\label{fig:sc_circuits}
\end{figure}

\paragraph{Bitstream generation.}
Stochastic number generation (SNG) is implemented in hardware as an integer comparator.
At each cycle $\ell$, a pseudo-random number generator (RNG) draws a uniform integer $r_\ell$ from the same $b$-bit range as the input, and the comparator emits $z_\ell=1$ if $r_\ell$ is below the input value, else $z_\ell=0$ (Figure~\ref{fig:sc_circuits}(a)).
For a signed integer $\hat{x}\in[-q_{\max},q_{\max}]$ with $q_{\max}=2^{b-1}-1$ encoding the bipolar value $v=\hat{x}/q_{\max}\in[-1,1]$, the input is first offset onto the unsigned RNG range as $\hat{x}+q_{\max}\in[0,\,2q_{\max}]$ before comparison, which gives $\Pr[z_\ell=1]=(v+1)/2$ as required by the bipolar encoding above.
Decoding inverts the encoding: from the count $c=\sum_\ell z_\ell$, the unbiased estimate of $v$ is $\tilde{v}=(2c-L)/L$.

The choice of RNG drives both the convergence of $\tilde{v}\to v$ at finite $L$ and the cross-correlation between any two streams produced for the same matrix multiplication; the latter is the principal source of SC error and is addressed by the C-BSG construction we adopt (Appendix~\ref{app:cbsg}).

\paragraph{SC matrix multiplication.}
SC enters the inference pipeline only at the integer matrix-multiply step.
We assume the inputs to a matrix multiplication $C=AB^\top$ with $A\in\mathbb{R}^{N\times D}$ and $B\in\mathbb{R}^{M\times D}$ have already been quantized by a standard PTQ pipeline~\citep{nagel2021ptq,frantar2023optq,chen2025qdit}: per-row symmetric scales $s^A_i=\max_d|A_{id}|/q_{\max}$ and $s^B_j=\max_d|B_{jd}|/q_{\max}$ produce $b$-bit signed integer matrices $\hat{A},\hat{B}$ with $\hat{A}_{id},\hat{B}_{jd}\in[-q_{\max},q_{\max}]$ such that $A_{id}\approx s^A_i\hat{A}_{id}$ and $B_{jd}\approx s^B_j\hat{B}_{jd}$.
In the subsequent integer matrix multiplication, an SC primitive replaces the multi-bit integer multiplier at each inner-dimension position.
Let $D$ denote the inner dimension shared by $\hat{A}$ and $\hat{B}$ and let $d\in\{1,\ldots,D\}$ index it.
Map each integer to a normalized bipolar value $u^A_{id}=\hat{A}_{id}/q_{\max}\in[-1,1]$, $u^B_{jd}=\hat{B}_{jd}/q_{\max}\in[-1,1]$, and encode each as a bipolar stream of length $L$.
For two such streams the bipolar product is captured by the bitwise XNOR gate: in expectation, the rescaled XNOR count over $L$ cycles equals $u^A_{id}\,u^B_{jd}$~\citep{alaghi2013survey} (the analogous unipolar AND construction is illustrated in Figure~\ref{fig:sc_circuits}(b)).
Summing over the inner dimension $d$ and rescaling by the per-row scales then gives the SC dot-product estimator
\begin{equation}
\widehat{C}_{ij} \;=\; s^A_i\, s^B_j\, q_{\max}^{2} \left[\,\frac{2}{L}\sum_{\ell=1}^{L}\sum_{d=1}^{D}\mathrm{XNOR}\!\bigl(z^A_{id,\ell},\,z^B_{jd,\ell}\bigr) \;-\; D\,\right],
\qquad \mathrm{XNOR}(a,b)\equiv 1-(a\oplus b),
\label{eq:sc_matmul}
\end{equation}
which is unbiased: $\mathbb{E}[\widehat{C}_{ij}]=C_{ij}$.
Throughout this paper, we generate the paired streams using the state-of-the-art \emph{conditional bitstream generation} (C-BSG) construction~\citep{ugemm,usystolic}, which enforces pairwise independence by construction so that \eqref{eq:sc_matmul} is bias-free at finite $L$ rather than stochastic; we defer the full derivation, including the formal stochastic cross-correlation definition, to Appendix~\ref{app:cbsg}.

\paragraph{Precision and early termination.}
The stream length $L$ is the precision knob.
Under the pairwise-independence assumption---enforced by C-BSG (Appendix~\ref{app:cbsg})---the per-pair XNOR estimator has variance $(1-(u^A_{id}u^B_{jd})^2)/L\le 1/L$~\citep{alaghi2013survey}, so the matmul estimator has worst-case standard deviation $\Theta\!\bigl(s^A_i s^B_j q_{\max}^{2}\sqrt{D/L}\bigr)$.
Crucially, $L$ can be \emph{any} positive integer, not only a power of two, so the same datapath exposes a dense, hardware-shared precision ladder; a scheduler can \emph{terminate early} for less salient outputs by stopping at $L'<L$, trading quality for proportional time and energy savings.
As a concrete example, an INT8 operand ($b=8$) recovers its full representable precision at $L=2^b=256$ cycles; truncating to $L'=192$ leaves $\log_2 L'\approx 7.58$ effective bits---a precision unreachable from a conventional integer datapath.

\subsection{Error Tolerance in Neural Networks}
\label{sec:nn_tolerance}

Neural networks routinely tolerate bounded numerical error from quantization, pruning, low-rank approximation, and noisy hardware, so their workloads are natural candidates for SC. 
To quantify the relevant tolerance, we replace each exact output cell $Y_{ij}=\langle A_{i,:},B_{j,:}\rangle$ with a Gaussian-perturbed surrogate $\widehat Y_{ij}=Y_{ij}(1+\alpha Z_{ij})$, $Z_{ij}\stackrel{\mathrm{i.i.d.}}{\sim}\mathcal{N}(0,1)$, with $\alpha$ calibrated as a function of stream length $L$ to match measured SC error on representative tensors. We apply this surrogate to all DiT-XL/2~\citep{peebles2023dit} matmuls.

\begin{center}
\begin{tabular}{cccccc}
\toprule
\makecell{stream length\\$L$ (sc\_len)}
&
\makecell{equiv.\ INT bits\\$\log_2 L$}
&
noise $\alpha$
&
latent MSE
&
image MSE
&
\makecell{image\\PSNR (dB)}
\\
\midrule
$32$  & $5.00$ & $0.0726$ & $3.2\!\times\!10^{-2}$ & $4.8\!\times\!10^{-3}$ & $23.2$ \\
$64$  & $6.00$ & $0.0363$ & $8.1\!\times\!10^{-3}$ & $1.3\!\times\!10^{-3}$ & $29.0$ \\
$128$ & $7.00$ & $0.0182$ & $3.8\!\times\!10^{-3}$ & $6.4\!\times\!10^{-4}$ & $32.0$ \\
$192$ & $7.58$ & $0.0121$ & $9.7\!\times\!10^{-4}$ & $2.0\!\times\!10^{-4}$ & $37.0$ \\
$256$ & $8.00$ & $0.0091$ & $7.9\!\times\!10^{-4}$ & $1.6\!\times\!10^{-4}$ & $37.9$ \\
\bottomrule
\end{tabular}
\end{center}

The sweep confirms that the neural workload has a usable fluctuation margin for SC: $L=256$ gives $37.9$\,dB PSNR, $L=192$ gives $37.0$\,dB, and $L=128$/$96$ remain near the quality range used in prior quantized diffusion evaluations~\citep{chen2025qdit,li2023qdiffusion}; quality drops at $L=64$ and degrades substantially for $L\le32$. This tolerance is the key hardware opportunity. Since SC exposes precision as the runtime stream length~$L$, including non-power-of-two operating points such as $L=96$ and $L=192$ without changing the circuit, we can spend longer streams only where the model is sensitive.

\section{Methodology}
\label{sec:method}

\begin{figure}[ht]
\centering
\includegraphics[width=1\linewidth]{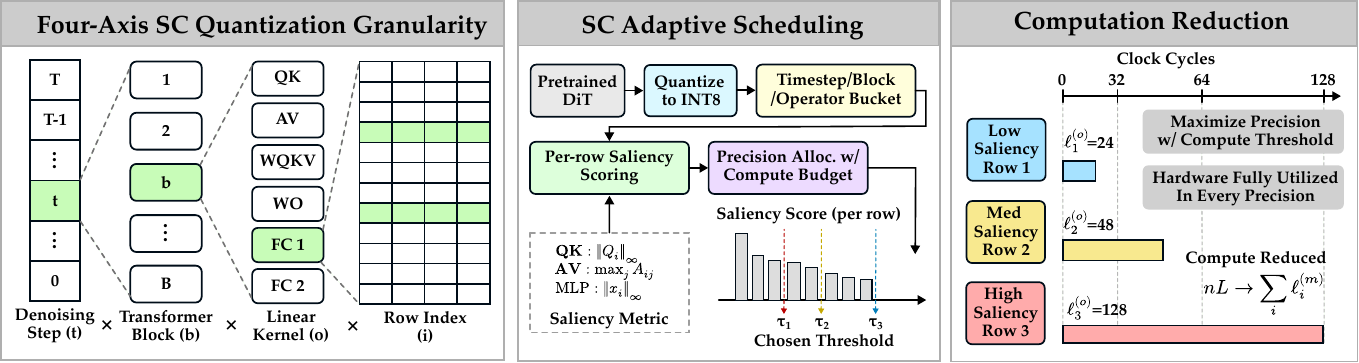}
\caption{Overview of our adaptive fine-grained quantization and scheduling flow.}
\label{fig:Flow}
\end{figure}

\subsection{Hardware Model}
\label{sec:hardware_model}

SC replaces a multi-bit multiplier with a one-bit gate (AND for unipolar, XNOR for bipolar) and pays precision temporally: a length-$L$ stream runs the same datapath for $L$ cycles.

\paragraph{Gate-level advantage.}
We synthesize four multipliers (SC vs.\ binary, signed vs.\ unsigned) in FreePDK45 \citep{FREEPDK} at 400\,MHz; Table~\ref{tab:hw_metric} reports area, power, and the per-operation energy reduction at $L=2^k$. The unipolar AND is $268\times$ smaller and $453\times$ lower-power than an unsigned INT8 multiplier, and the bipolar XNOR is $156\times$ smaller and $399\times$ lower-power than its signed counterpart. After the $L$-cycle penalty, SC is still $1.6$--$1.8\times$ lower-energy at $k{=}8$ and $6$--$7\times$ at $k{=}6$. The area gap also buys throughput: the $156\times$-smaller XNOR can support $\sim$$128\times$ more lanes at iso-area, offsetting the $L\times$ latency factor.

\begin{table}[h]
\centering
\small
\caption{SC vs.\ binary multiplier comparison at 400\,MHz.}
\label{tab:hw_metric}
\begin{tabular}{lcccccccc}
\toprule
& \multicolumn{2}{c}{SC gate} & \multicolumn{2}{c}{INT8 mult.} &
  \multicolumn{4}{c}{Energy/Op Reduction} \\
\cmidrule(lr){2-3}\cmidrule(lr){4-5}\cmidrule(lr){6-9}
Pair & Area & Power & Area & Power & $k=8$ & $k=7$ & $k=6$ & $k=5$ \\
& ($\mu$m$^2$) & ($\mu$W) & ($\mu$m$^2$) & ($\mu$W) & & & & \\
\midrule
\textsc{And}/\textsc{UMul}
  & 1.06 & 0.37 & 285.4 & 168.7
  & $1.77\times$ & $3.54\times$ & $7.08\times$ & $14.2\times$ \\
\textsc{Xnor}/\textsc{Mul}
  & 1.60 & 0.63 & 249.2 & 249.7
  & $1.56\times$ & $3.13\times$ & $6.25\times$ & $12.5\times$ \\
\bottomrule
\end{tabular}
\end{table}

\paragraph{Full-accelerator evidence.}
The gate-level gap survives bit-stream generation, accumulation, control, and data movement at chip scale: Cambricon-U reports $2.1\times$ lower energy and $2.2\times$ smaller area than an INT8 baseline on MLPerf-Tiny, and $1.22$--$1.47\times$ energy-efficiency gains across MLPerf-Tiny and four ImageNet-scale CNNs without accuracy loss; uGEMM saves $72\%$ of cycles via early termination at higher MLP accuracy than prior unary designs.

\paragraph{From compute cycles to energy and latency.}
For matmul $m$ with per-row stream length $\ell_i^{(m)}$, we take SC compute cost as $C_{\mathrm{SC}} \propto \sum_m \sum_i \ell_i^{(m)}$. Because every stream cycle drives the same gate, energy is linear in $C_{\mathrm{SC}}$, and latency is linear at fixed lane count. Two schedules with matched $\sum \ell_i$ therefore have matched energy and latency, and reducing the average stream length from $L$ to $\bar\ell$ yields $L/\bar\ell$ savings on both axes (e.g., $\bar\ell{=}128$ or $64$ at $L{=}256$ gives $2\times$ or $4\times$). Section~\ref{sec:mixed_precision} spends this budget by shortening streams on less sensitive rows.

\subsection{GPU Kernel for SC Matrix Multiplication}
\label{sec:gpu_kernel}

\paragraph{GPU execution path.}
Existing CPU SC simulators are too slow to insert into every transformer matmul, while the C-BSG datapath is a regular tiled AND-and-popcount computation that fits GPU execution. We implement the uSystolic-style C-BSG temporal-unary multiplier from Section~\ref{sec:sc_basics} as fused Triton kernels~\cite{triton}; in our benchmark, this path is about $95{,}000\times$ faster than direct CPU SC simulation. The kernel treats stream length as data, so per-row length vectors can be changed at runtime without recompilation.

\paragraph{Datapath: cumulative-indicator C-BSG.}
Each matmul begins with fused quantization: operands are mapped to the int8 grid ($q_{\max}=127$), converted to one of $V=257$ comparison boundaries, and emitted with row scale and sign in one pass. The C-BSG stream is represented by a precomputed cumulative-indicator table $K[b,\ell]=\mathbf{1}[r_\ell\le b]$ over Sobol-8 values~\citep{sobol1967,joekuo2008sobol,scgen}; the compact kernel indexes $K$ by the weight and activation boundary tensors and accumulates the bipolar SC estimator as tiled AND-and-popcounts over $D{\times}L$ binary comparisons. The table is built once per layer at $L_{\max}=2^8$, cached on device, and reused by taking prefixes for shorter streams. This same launch receives a row-length vector $\boldsymbol{\ell}\in\mathbb{Z}_{\ge 0}^{N}$, so row $i$ stops at $\boldsymbol{\ell}_i$ and different rows can run at different effective precisions for Section~\ref{sec:mixed_precision}.

\subsection{Sensitivity Analysis}
\label{sec:sensitivity_analysis}

\paragraph{Probing where SC error matters.}
SC trades a multi-bit multiplier for one-bit logic, but its arithmetic is only statistically correct: the C-BSG kernel of Section~\ref{sec:gpu_kernel} is unbiased, yet matching a fixed-point multiplier's variance can require stream lengths in the hundreds or thousands~\citep{wu2022usystolic, guo2023cambricon}. Paying this cost uniformly is wasteful---prior mixed-precision work shows numerical sensitivity is highly concentrated, with a small subset of layers and matmul classes dominating the error budget~\citep{mixdq, mixdit}. We therefore probe \emph{where} SC precision must be spent. For each operator class
$o\in\mathcal{O}=\{\text{qk},\text{av},\text{qkv\_proj},\text{out\_proj},\text{fc1},\text{fc2}\}$
and block $b\in\{0,\ldots,B-1\}$, we route only that matmul through C-BSG at stream length $L$, leaving all others in full precision, and measure
\begin{equation}
\mathrm{S}(o,b,L) \;=\; \mathbb{E}_{x\in\mathcal{D}}\,\bigl\|\,h^\star(x) - \widehat h^{(o,b,L)}(x)\,\bigr\|_2,
\label{eq:sensitivity_def}
\end{equation}
where $h^\star$ and $\widehat h^{(o,b,L)}$ are the full-precision and single-SC backbone feature tensors on a small calibration set $\mathcal{D}$.


\begin{figure}[ht]
\centering
\includegraphics[width=\linewidth]{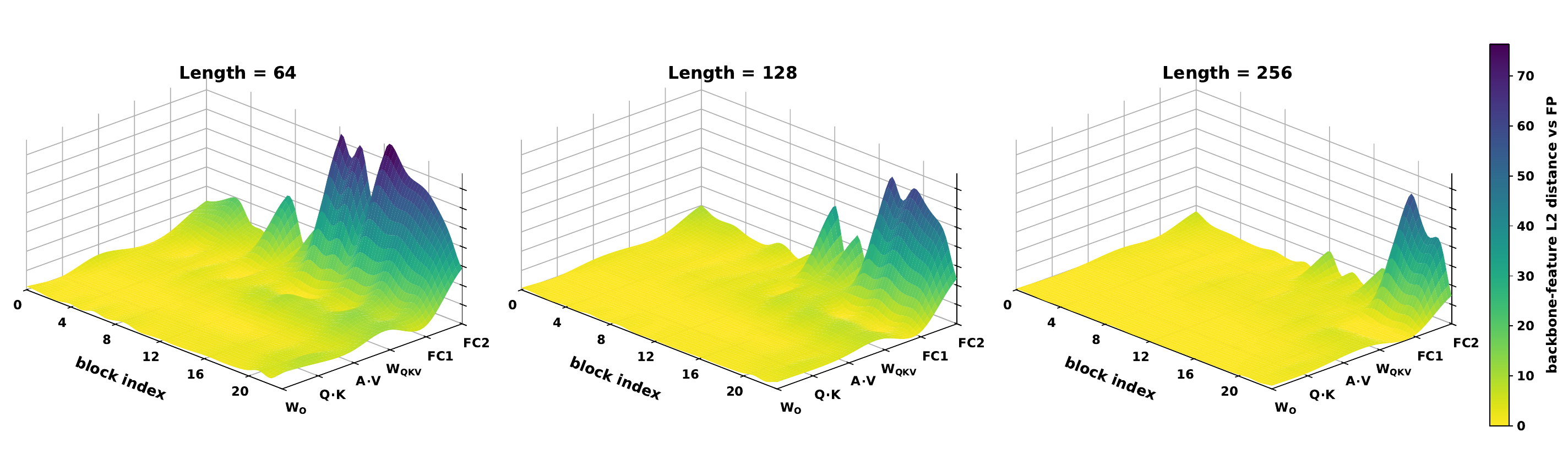}
\caption{SC sensitivity is highly non-uniform across operator class, block depth, and stream length: replacing a single $(o,b)$ matmul with SC perturbs the final ViT-L/14 feature far more for late \textsc{FC2} blocks than for any other tile, motivating the mixed-precision policy of Section~4.4.}
\label{fig:sensitivity_surface_3d}
\end{figure}

\paragraph{Case study on ViT-L/14 + ImageNet.}
We instantiate the probe on a pretrained DINOv2 ViT-L/14 backbone ($B=24$, $|\mathcal{O}|=6$) with $|\mathcal{D}|=100$ ImageNet images at three stream lengths $L\in\{64,128,256\}$, yielding the sweep in Figure~\ref{fig:sensitivity_surface_3d}. 
The results reveal two patterns. First, \textsc{FC2} dominates the high-sensitivity region across stream lengths, indicating SC error is far from uniform across operator classes. Second, within \textsc{FC2} sensitivity concentrates in later blocks: at $L=64$, late-\textsc{FC2} accounts for $19$ of the top-$20$ most sensitive $(o,b)$ tiles. These motivate reserving binary fallback for the most sensitive tiles and using the mixed-precision policy of Section~\ref{sec:mixed_precision} to allocate longer SC streams to sensitive operators and shorter streams to less sensitive ones.

\subsection{Per-Row Mixed-Precision SC}
\label{sec:mixed_precision}

\paragraph{Calibration mode.}
Section~\ref{sec:gpu_kernel} exposes stream length as a per-row runtime argument, so mixed precision reduces to choosing the stream length assigned to each output row of each SC matmul.
We make this choice through an offline \emph{calibration mode}: for each target average stream length $\bar L$, the calibrator runs a small set of FP-teacher trajectories, replays each selected matmul with uniform SC at several candidate stream lengths, and records the row-wise deviation.
The result is a JSON table of metric thresholds.
Inference only evaluates the same lightweight row metrics, looks up the calibrated thresholds, and passes the resulting row-length vector to the SC kernel.

\paragraph{Row metrics.}
For each calibrated operator, the row metric is chosen from tensors already present in the forward pass.
Following AWQ~\citep{lin2024awq} and SmoothQuant~\citep{xiao2023smoothquant}, for \textsc{qk} we use the row-wise query magnitude, $g_i^{\textsc{qk}}=\|Q_i\|_\infty$.
For \textsc{av}, we use the peak attention probability, $g_i^{\textsc{av}}=\max_j A_{ij}$.
For \textsc{input\_proj}, \textsc{proj}, \textsc{mlp\_fc1}, and \textsc{mlp\_fc2}, we use the row-wise activation magnitude, $g_i=\|X_i\|_\infty$.
We use the maximum magnitude $a_{\max}$ rather than an average because the per-row SC quantizer is scaled by the row range: the largest entry sets the comparison boundary scale and also bounds the largest contribution any coordinate can make to the dot-product error.
Thus $a_{\max}$ is a cheap upper-envelope proxy for row dynamic range, and it is already available from the fused per-row scale extraction.
Within each operator call, metrics are normalized to $[0,1]$ as
\[
\tilde g_i=\frac{g_i-\min_j g_j}{\max_j g_j-\min_j g_j}.
\]
Rows with larger $\tilde g_i$ are treated as more precision-sensitive and therefore receive longer streams after thresholding.
\paragraph{Threshold fitting.}
Let $L_1>L_2>\cdots>L_K$ be the stream-length ladder supplied to the calibrator.
For each operator, timestep bucket, and layer-depth bucket, the calibrator measures row-wise cosine error against the FP teacher at every $L_k$ and solves a budgeted Lagrangian assignment:
\[
k_i(\lambda)=\arg\min_k \bigl(e_{ik}+\lambda L_k\bigr),
\]
where $e_{ik}$ is the measured FP-vs-uniform-SC error.
Binary search on $\lambda$ matches the target budget, and the resulting level counts are converted into non-uniform thresholds
\[
1\ge\tau_1\ge\tau_2\ge\cdots\ge\tau_{K-1}\ge 0.
\]
At inference, row $i$ receives the first level whose threshold it exceeds, $k^\star(i)=\min\{k:\tilde g_i\ge\tau_k\}$, or the shortest stream if it falls below all thresholds; the kernel then uses $\ell_i=L_{k^\star(i)}$.
The JSON table is keyed by operator, timestep bucket, and layer bucket, with operator-level fallback for sparse buckets.

\section{Experiments}
\label{sec:experiments}

\subsection{Classification}
\label{sec:exp_classification}

\paragraph{Setup.}
We evaluate DINOv2 ViT-L/14~\citep{oquab2023dinov2} on full ImageNet-1k val ($N{=}50{,}000$), comparing SC---with and without our residual \emph{compensation} (Appendix~\ref{app:qwt_sc})---against the \emph{QwT}~\citep{qwt} baseline (INT quantization~\citep{li2023repq} with closed-form residual compensation). For SC we further compare uniform-$L$ against the per-row mixed-precision policy (\emph{MP}) of Section~\ref{sec:mixed_precision} at matched bit budget $\sum_i \ell_i$; all SC runs keep the $30$ most-sensitive (op, block) cells from the Section~\ref{sec:sensitivity_analysis} profile in FP. Bit budgets follow the W/A convention of \citet{chen2025qdit}.

\begin{table}[ht]
\centering
\caption{\textbf{ImageNet-1k classification (full val, $N{=}50{,}000$).} \emph{W/A} is weight/activation bits: the underlying INT setting for QwT; for SC, $W{=}A{=}\log_2 \bar L$ with $\bar L$ the average per-row stream length.}
\label{tab:cls_imagenet}
\small
\setlength{\tabcolsep}{6pt}
\renewcommand{\arraystretch}{1.05}
\begin{tabular}{l c c c c}
\toprule
Method & W/A & top-1 (\%) & top-5 (\%) & $\Delta$ vs FP \\
\midrule
FP (DINOv2 ViT-L/14)             & $16/16$              & 86.40 & 97.65 & $\;\;\;0.00$ \\
\midrule
QwT                              & $8/8$                & 84.57 & 96.56 & $-1.83$ \\
SC uniform                       & $8/8$                & 85.57 & 97.40 & $-0.83$ \\
SC uniform + compensation        & $8/8$                & 85.70 & 97.47 & $-0.70$ \\
SC uniform                       & $\log_2\!192/\log_2\!192$      & 84.67 & 97.11 & $-1.73$ \\
SC uniform + compensation        & $\log_2\!192/\log_2\!192$      & 85.42 & 97.43 & $-0.98$ \\
SC MP                            & $\log_2\!192/\log_2\!192$      & 85.07 & 97.23 & $-1.33$ \\
SC MP + compensation             & $\log_2\!192/\log_2\!192$      & 85.47 & 97.40 & $-0.93$ \\
\midrule
QwT                              & $4/8$                & 83.48 & 96.37 & $-2.92$ \\
SC uniform                       & $7/7$                & 78.94 & 94.61 & $-7.46$ \\
SC uniform + compensation        & $7/7$                & 82.87 & 96.46 & $-3.53$ \\
SC MP                            & $7/7$                & 79.84 & 94.90 & $-6.56$ \\
SC MP + compensation             & $7/7$                & 83.26 & 96.45 & $-3.14$ \\
\bottomrule
\end{tabular}
\end{table}

\paragraph{Results.}
SC + compensation tracks QwT at every matched bit budget in Table~\ref{tab:cls_imagenet}. MP beats uniform-$L$ at matched average budget by recovering the asymmetry Section~\ref{sec:sensitivity_analysis} flagged---raising \textsc{mlp\_fc2} to $L{=}192$ in every active block at $\bar L{=}128$ ($L{=}256$ in $40\%$ of blocks at $\bar L{=}192$) and letting the cheap projections absorb $L{\in}\{64,96\}$ in $\sim$$36\%$ of blocks. Compensation lifts SC by $+0.13$\,pt at $L{=}256$ up to $+3.93$\,pt at $L{=}128$; the closed-form fit of \citet{qwt} only needs errors deterministic in $(w, x)$, which holds for SC because our seed-fixed Sobol streams fix the per-element error at calibration time, exactly as it holds for INT under deterministic rounding (Appendix~\ref{app:qwt_sc}).

\subsection{Object Detection and Instance Segmentation}
\label{sec:exp_detection_segmentation}

\paragraph{Setup.}
We evaluate EVA-01 ViT-g + cascade Mask R-CNN~\citep{fang2023eva,li2022exploring,he2017mask} on COCO val ($N{=}5000$, square-padded to $1024^2$, soft-NMS, BEIT pos.\ interp.). SC rows skip the $30$ most-sensitive of $240$ (op, block) tiles to FP (Section~\ref{sec:sensitivity_analysis}); MP allocates within the active set per Section~\ref{sec:mixed_precision}. We compare SC (uniform-$L$, MP, $\pm$ compensation) against QwT-INT~\citep{qwt} at matched $W{=}A$.

\begin{table}[ht]
\centering
\caption{\textbf{COCO val object detection \& instance segmentation, EVA-01 ViT-g + cascade Mask R-CNN ($N{=}5000$, $1024^2$).} \emph{W/A} as in Table~\ref{tab:cls_imagenet}.}
\label{tab:det_coco}
\small
\setlength{\tabcolsep}{4pt}
\renewcommand{\arraystretch}{1.05}
\begin{tabular}{l c c c c}
\toprule
Method & W/A & AP$_b$ & AP$_m$ & $\Delta_b/\Delta_m$ \\
\midrule
FP (EVA-01 ViT-g)                & $16/16$                        & 62.06 & 52.39 & $\;\;\;0.00/\;\;0.00$ \\
\midrule
QwT                              & $8/8$                          & 61.20 & 51.59 & $-0.86/-0.80$ \\
SC uniform                       & $8/8$                          & 61.98 & 52.18 & $-0.08/-0.21$ \\
SC uniform                       & $\log_2\!192/\log_2\!192$      & 61.69 & 51.85 & $-0.37/-0.54$ \\
SC uniform + compensation        & $\log_2\!192/\log_2\!192$      & 61.82 & 52.06 & $-0.24/-0.33$ \\
\midrule
QwT                              & $7/7$                          & 61.13 & 51.51 & $-0.93/-0.88$ \\
SC uniform                       & $7/7$                          & 58.84 & 48.68 & $-3.22/-3.71$ \\
SC uniform + compensation        & $7/7$                          & 59.72 & 49.81 & $-2.34/-2.58$ \\
SC MP                            & $7/7$                          & 60.45 & 50.76 & $-1.61/-1.63$ \\
\bottomrule
\end{tabular}
\end{table}

\paragraph{Results.}
SC uniform at $8/8$ is essentially FP-equivalent on boxes
($61.98$ vs.\ $62.06$ AP$_b$) and clearly outperforms QwT $8/8$
($61.20$). At the $\log_2\!192$ tier, compensation gives a small but
consistent gain over raw uniform SC ($61.82/52.06$ vs.\ $61.69/51.85$
AP$_b$/AP$_m$). The harder regime is $7/7$: uniform SC drops to
$58.84/48.68$, while compensation recovers $+0.88/+1.13$ AP and MP
recovers $+1.61/+2.08$ AP, making MP the stronger lever at this budget.


\subsection{Image Generation}
\label{sec:exp_image_generation}

\paragraph{Setup.}
We evaluate SC on class-conditional ImageNet generation with DiT-XL/2 at $256{\times}256$ resolution~\citep{peebles2023dit}.
All attention and MLP matrix multiplications are routed through the GPU C-BSG kernel of Section~\ref{sec:gpu_kernel}; the quantization and stream-length schedules follow the same bit-budget convention used in Sections~\ref{sec:exp_classification} and~\ref{sec:mixed_precision}.
All rows use the same sampler with $50$ denoising timesteps and classifier-free guidance scale $\mathrm{CFG}{=}4.0$.
We generate $10{,}000$ images in total, with $10$ samples for each of the $1000$ ImageNet classes.
We report IS, sFID, KID, and FID against the ImageNet validation reference set, and include the realized average stream length to make the compute budget explicit.

\begin{table}[t]
\centering
\caption{\textbf{ImageNet class-conditional generation with DiT-XL/2, $256{\times}256$.} All rows use the same sampler, $50$ denoising timesteps, $\mathrm{CFG}{=}4.0$, and $10{,}000$ generated images ($10$ per ImageNet class). Schedule records the SC length setting, $\bar L$ is the realized average stream length over SC-executed matmuls, and KID is reported as $10^3{\times}\mathrm{KID}$.}
\label{tab:image_generation}
\begin{tabular}{lcccccc}
\toprule
Method  & $\bar L$ & IS$\uparrow$ & sFID$\downarrow$ & KID$\times10^3\downarrow$ & FID$\downarrow$ \\
\midrule
FP & -- & $285.95$ & $24.75$ & $ 13.47{\pm}1.24$ & $24.81$ \\
SC MP & 64 & $279.76$ & $19.93$ & $9.7717 {\pm}1.0457 $ & $17.33$ \\
SC uniform  & 96 & $284.62$ & $20.91$ & $11.1134{\pm}1.1318$ & $18.47$ \\
\bottomrule
\end{tabular}
\end{table}

\paragraph{Result.}
Table~\ref{tab:image_generation} summarizes the image-generation results.
The comparison is intended to test whether the SC schedules that preserve recognition accuracy also maintain generative sample quality at matched average stream length.
SC matching or slightly improving over FP on FID is reasonable in this setting: diffusion sampling is explicitly noise-driven, and the iterative denoising process is robust to small arithmetic perturbations, so moderate SC noise need not translate into worse sample-level statistics.
Qualitative samples are shown in Appendix~\ref{app:image_samples}, Figure~\ref{fig:image_generation_samples}.

\subsection{Visual World-Model Planning}
\label{sec:exp_world_model}

\paragraph{Setup.}
We test SC's feasibility in a closed-loop visual planner, where backbone errors compound through a learned dynamics model over long rollouts.
We SC-quantize the full Dino-WM~\citep{zhou2024dino} stack on \texttt{wall\_single} (2D goal-reaching): a pretrained DINOv2 ViT-S/14 encoder and a 6-block latent predictor. We keep the
$12/60$ most-sensitive encoder tiles and $7/36$ predictor tiles in FP and SC-quantize the rest. Planning uses CEM-MPC with
the same budget for all rows of Table~\ref{tab:wm_wall}
($n_{\mathrm{evals}}{=}30$ episodes, $\mathrm{max\_iter}{=}20$, $900$ samples,
seed-fixed); SR resolution is $1/30 \!\approx\! 0.033$.

\begin{table}[ht]
\centering
\caption{\textbf{Visual planning success rate on Dino-WM \texttt{wall\_single}, matched CEM budget.} SR is the fraction of $30$ episodes solved within $20$ MPC steps; bit budget follows Table~\ref{tab:cls_imagenet}.}
\label{tab:wm_wall}
\small
\setlength{\tabcolsep}{6pt}
\renewcommand{\arraystretch}{1.05}
\begin{tabular}{l c c c c}
\toprule
Method & W/A & SR & Plateau iter & $\Delta$ vs FP \\
\midrule
FP (Dino-WM, ViT-S/14 enc.\ + 6-block pred.) & $16/16$            & $0.833$ & $5$  & $\;\;\;0.000$ \\
\midrule
SC uniform              & $8/8$                          & $0.833$ & $3$  & $\;\;\;0.000$ \\
SC uniform              & $\log_2\!192/\log_2\!192$      & $0.833$ & $3$  & $\;\;\;0.000$ \\
SC MP                   & $\log_2\!192/\log_2\!192$      & $0.933$ & $16$ & $\;\;\;{+}0.100$ \\
\midrule
SC uniform              & $7/7$                          & $0.800$ & $16$ & $-0.033$ \\
SC MP                   & $7/7$                          & $0.833$ & $13$ & $\;\;\;0.000$ \\
\bottomrule
\end{tabular}
\end{table}

\paragraph{SC matches FP across both budgets, while MP provides low-precision headroom.}
At the $\sim$8-bit tier, all SC configurations reach FP's $0.833$ SR plateau,
and both uniform variants reach their plateau two CEM iterations earlier than FP.
SC MP at $\log_2\!192$ reaches $0.933$, solving two additional episodes within
the 20-iteration horizon. We do not over-interpret this gain ($\pm 2$ episodes
per seed at $1/30$ SR resolution from sampling alone), but it is consistent with
a small structured effect of MP's per-row stream-length perturbations on CEM
elite selection. At the 7-bit tier, uniform SC falls one episode short
($0.800$), while SC MP at the same average budget recovers FP-equivalent SR
($0.833$). This mirrors Table~\ref{tab:cls_imagenet}: per-row MP recovers the
headroom that uniform SC loses below the safe precision regime, now in a
closed-loop planning task rather than only in a static classifier readout.

\section{Discussion}
\label{sec:discussion}

\paragraph{Limitations.}
Our evaluation covers only a limited set of configurations: for each task we use
representative backbones and a small number of precision budgets, rather than an
exhaustive sweep over models, seeds, and more aggressive low-bit settings. Thus,
our results establish SC's feasibility in the evaluated regimes, but do not yet
show how robust the conclusions are across a broader configuration space.
In addition, the hardware benefits are estimated from analytical and gate-level models rather than measured end-to-end silicon.

\paragraph{Broader impacts.}
This work supports broader exploration of SC for modern AI inference. Since specialized hardware requires
substantial design effort, software-level feasibility evidence is important for
deciding which arithmetic paradigms are worth pursuing in silicon. Our results
suggest that dense, temporally controlled SC precision is a promising direction
for future accelerator design.

\section{Conclusion}
\label{sec:conclusion}
This paper argues that stochastic computing is best viewed as a \emph{dense adaptive quantizer} for modern transformer inference: precision is controlled by stream length rather than constrained to a small set of fixed word sizes.
Empirically, we show that SC retains near-lossless accuracy on transformer-based vision tasks---image classification, object detection and instance segmentation, class-conditional image generation, and visual world-model planning.
To make this concrete at scale, we contribute a fused GPU library for SC matrix multiplication on modern vision transformers and a per-row mixed-precision algorithm that turns row-level sensitivity into a calibrated stream-length budget without retraining.
Together, these results position SC as a viable substrate for future adaptive-precision accelerators, where a simple shared datapath can expose a much denser precision space than conventional fixed-format arithmetic.

\bibliographystyle{plainnat}
\bibliography{references}

\appendix
\section{Supplementary Material}
\subsection{Correlation and the C-BSG Construction}
\label{app:scc}
\label{app:cbsg}

This appendix expands the brief reference to conditional bitstream generation
(C-BSG) in Section~\ref{sec:sc_basics}: it gives the formal definition of
stochastic cross-correlation, motivates why a non-zero correlation biases the
SC estimator in \eqref{eq:sc_matmul}, and presents the UGEMM/C-BSG construction
that we use throughout this paper to enforce zero correlation by construction.

The independence assumption behind \eqref{eq:sc_matmul} is the principal source of SC error in practice.
Following~\citet{alaghi2013survey,ugemm}, let $a$, $b$, $c$, $d$ count the cycles at which the bit pair $(z^A_\ell,z^B_\ell)$ equals $(1,1)$, $(1,0)$, $(0,1)$, $(0,0)$ respectively (with $a+b+c+d=L$); the \emph{stochastic cross-correlation} $\mathrm{SCC}(Z^A,Z^B)\in[-1,+1]$ is a normalized form of $ad-bc$ that vanishes exactly when the two streams are pairwise independent, with full piecewise definition
\begin{equation}
\mathrm{SCC}(Z^A,Z^B) \;=\; \frac{ad-bc}{\,D\,},
\qquad
D \;=\;
\begin{cases}
L\,\min(a+b,\,a+c) - (a+b)(a+c), & \text{if } ad>bc,\\[2pt]
(a+b)(a+c) - L\,\max(a-d,\,0), & \text{otherwise}.
\end{cases}
\label{eq:scc_full}
\end{equation}
The two-case denominator normalizes $ad-bc$ by its maximum attainable absolute value subject to the marginals $(a+b)$ and $(a+c)$, so that $\mathrm{SCC}=\pm 1$ is reached exactly when $Z^A$ and $Z^B$ are perfectly correlated or anti-correlated.

Pseudo-random number generators such as LFSRs attain the standard $\mathcal{O}(L^{-1/2})$ decoding rate but offer no cross-correlation guarantees at finite $L$.
We instead use scrambled Sobol low-discrepancy sequences~\citep{sobol1967,joekuo2008sobol}: deterministic sequences in $[0,1)^d$ constructed by XOR-ing bit-indexed direction numbers, with $L^\infty$-discrepancy $\mathcal{O}(L^{-1}\log^d L)$ that is asymptotically tighter than the $\mathcal{O}(L^{-1/2})$ rate of i.i.d.\ samples~\citep{niederreiter1992quasi}.
Even so, Sobol RNGs reduce $|\mathrm{SCC}|$ but do not eliminate it at finite $L$, and the residual non-zero correlation biases the estimator of \eqref{eq:sc_matmul}.

We address this gap with the UGEMM~\citep{ugemm,usystolic} construction.
For two bipolar streams $z^A,z^B$ encoding $u^A,u^B\in[-1,1]$, the output bit value $z_\mathrm{out}=\mathrm{XNOR}(z^A,z^B)$ that decodes to $u^A u^B$ requires the output probability to satisfy~\citep{ugemm}
\begin{equation}
\Pr[z_\mathrm{out}\!=\!1] \;=\; \Pr[z^A\!=\!1]\,\Pr[z^B\!=\!1] \;+\; \Pr[z^A\!=\!0]\,\Pr[z^B\!=\!0],
\label{eq:bipolar_product}
\end{equation}
i.e., a function of the input \emph{marginals}.
A bitwise XNOR over two streams, however, fires whenever the two bits agree, so its output probability is the \emph{joint} probability of agreement, which factors via the chain rule as~\citep{ugemm}
\begin{equation}
\begin{aligned}
\Pr[z_\mathrm{out}\!=\!1]
&\;=\; \Pr[z^A\!=\!1,\,z^B\!=\!1] \;+\; \Pr[z^A\!=\!0,\,z^B\!=\!0]\\
&\;=\; \Pr[z^A\!=\!1]\,\Pr[z^B\!=\!1\mid z^A\!=\!1] \;+\; \Pr[z^A\!=\!0]\,\Pr[z^B\!=\!0\mid z^A\!=\!0],
\end{aligned}
\label{eq:bipolar_joint}
\end{equation}
which matches \eqref{eq:bipolar_product} only when both conditional probabilities equal the corresponding marginal:
\begin{equation}
\Pr[z^B\!=\!1\mid z^A\!=\!1]\;=\;\Pr[z^B\!=\!1], \qquad \Pr[z^B\!=\!0\mid z^A\!=\!0]\;=\;\Pr[z^B\!=\!0].
\label{eq:cbsg}
\end{equation}
\emph{Conditional bitstream generation} (C-BSG)~\citep{ugemm,usystolic} enforces \eqref{eq:cbsg} by construction (Figure~\ref{fig:sc_circuits}(c)): rather than generating $Z^A$ and $Z^B$ from two independent RNGs, the second stream is produced by an RNG whose update is gated by the first stream, so that the conditional distribution of $z^B$ given the value of $z^A$ is forced to equal its marginal.
Substituting \eqref{eq:cbsg} into the bit-pair counts above yields $ad-bc=0$ and hence $\mathrm{SCC}=0$, so the XNOR-based bipolar multiplier becomes \emph{bias-free at finite $L$} rather than stochastic.

\subsection{Residual Compensation in the SC Datapath}
\label{app:qwt_sc}

\paragraph{The QwT recipe we adapt.}
QwT~\citep{qwt} attaches one lightweight residual \emph{compensation} module to each transformer block of an already-quantized network. For block $b$ with quantized output $y^{Z}_b(x)=l^{Z}_b(x)$ and full-precision teacher output $y_b(x)=l_b(x)$, the module
\begin{equation}
\widehat y_b(x) \;=\; y^{Z}_b(x) \;+\; W_b\, x \;+\; b_b
\label{eq:qwt_block}
\end{equation}
adds an affine correction trained to minimize the per-block residual $\sum_{x\in\mathcal{C}}\,\|\,y_b(x)-y^{Z}_b(x)-W_b x-b_b\,\|_2^2$ on a small calibration set $\mathcal{C}$ of $\sim$$512$ ImageNet training images.
The minimizer is the ordinary least-squares solution $W_b=(Y_b-Y^{Z}_b)\,X_b^{\top}(X_b X_b^{\top})^{-1}$, with the bias absorbed by augmenting $X_b$ with a row of ones, so calibration is one closed-form solve per block and runs in minutes.
By construction $W_b{=}0$ recovers the uncompensated network, which makes QwT safe to attach unconditionally on the INT side; \citet{qwt} report up to $+8$\,pt at W4A4 RepQ-ViT and small but consistent gains at higher bit-widths.

\paragraph{Our SC-side compensation recipe.}
We adopt the same per-block, residual structure as~\eqref{eq:qwt_block}, with three modifications motivated by the SC residual being stochastic-but-deterministic (next paragraph) rather than a per-tensor integer round.
First, we use $|\mathcal{C}|{=}1024$ ImageNet train images and a ridge-regularized solve $W_b=(Y_b{-}Y^{Z}_b)\,X_b^{\top}(X_b X_b^{\top}+\lambda I)^{-1}$ with $\lambda{=}10^{-4}$, which we found necessary to keep $W_b$ well-conditioned in low-$L$ blocks where $X^{Z}_b$ has reduced effective rank.
Second, we fit and store $W_b$ \emph{per head} ($16$ heads $\times$ $d_{\text{head}}{=}64$) rather than as a single $d{\times}d$ matrix, which reduces parameter count and lets the compensator follow the head-aligned SC kernels of Section~\ref{sec:gpu_kernel}.
Third, we gate admission of $W_b$ into the inference path by a \emph{disjoint cross-seed cosine test}: we fit two independent solutions $W^{(A)}_b, W^{(B)}_b$ on disjoint $512$-image calibration halves and admit $W_b=\tfrac12(W^{(A)}_b+W^{(B)}_b)$ only when $\cos\!\bigl(W^{(A)}_b, W^{(B)}_b\bigr)>\tau$ ($\tau{=}0.5$ for blocks $0{-}22$, $\tau{=}0.8$ for the final block).
Failed gates leave the block uncompensated.
The gate exists because the SC residual fluctuates at finite calibration size, and an admitted $W_b$ that happens to fit calibration noise can degrade rather than improve held-out accuracy; the cross-seed test rejects exactly those overfit fits.
Empirically the gate admits $22/24$ blocks at $\bar L{\ge}128$ on ViT-L/14 (Section~\ref{sec:exp_classification}, Table~\ref{tab:cls_imagenet}).

\paragraph{The compensator is itself an SC matmul.}
A subtle but essential property of our pipeline is that the residual matmul $W_b x$ in~\eqref{eq:qwt_block} is \emph{not} an FP linear layer; it is a second SC dot-product evaluated on the C-BSG kernel of Section~\ref{sec:gpu_kernel} at uniform stream length $L_{\text{comp}}{=}256$ ($p{=}8$, bipolar polarity).
Calibration solves for the FP-valued $W_b$ in the closed form above, then materializes $W_b$ as the weight tensor of an SC linear layer that consumes the block input $x$ via the same fused per-row quantization kernel and the same cumulative-indicator $K$-table cache that the block's main matmuls use.
The whole inference path therefore stays inside the SC datapath: there is no point at which the network reintroduces an FP multiplier to ``rescue'' the quantized output, and the cost of a compensated block is one main-path SC matmul plus one auxiliary $d{\times}d$ SC matmul at fixed $L_{\text{comp}}{=}256$.
This is the property that lets the discussion of Section~\ref{sec:exp_classification} treat compensation as a pure precision/accuracy lever rather than as a hybrid INT--FP correction.

\paragraph{Deterministic SC noise lets $W_b$ \emph{absorb} a fixed realization.}
The C-BSG kernel of Section~\ref{sec:gpu_kernel} is stochastic in name only: its randomness is a length-$L_{\max}$ Sobol low-discrepancy sequence~\citep{sobol1967,joekuo2008sobol} indexed by a configuration (dim, seed, polarity) that is fixed at model load and reused on every forward pass.
Consequently the per-element SC error $\widehat{\langle w,x\rangle}-\langle w,x\rangle$ is a deterministic function of $(w,x)$ for a given configuration---it is ``shot noise'' only across hypothetical Sobol seeds, not across forward passes.
We exploit this by using the \emph{same} Sobol configuration during compensation calibration and during inference, so the residual $Y_b-Y^{Z}_b$ that $W_b$ is fit against is exactly the residual that $W_b$ encounters at test time.
In effect, $W_b$ does not need to ``average out'' SC noise; it absorbs a fixed deterministic perturbation, and any structure in that perturbation that is approximately linear in $x$ is fully captured by the closed-form fit.
This mirrors the interpretation \citet{qwt} give for INT-side QwT: where the INT residual is a deterministic, scale-induced rounding map, the SC residual is a deterministic, Sobol-induced kernel-evaluation map, and OLS is well-posed against either.

\paragraph{Scope: feasibility, not formal analysis.}
We frame this section strictly as a \emph{feasibility} demonstration: a closed-form fit, a cross-seed admission gate, and an SC-realized compensator together form a plausible recipe, and the empirical results of Section~\ref{sec:exp_classification} (within $1$\,pt of FP at $\bar L{\ge}192$, within $3.2$\,pt at $\bar L{=}128$) suggest the recipe is sound at the budgets we evaluate.
We do \emph{not} establish here (i)~that the deterministic-Sobol fixed-point characterization of $W_b$ is unique or stable across changes of Sobol configuration, (ii)~conditions under which the cross-seed cosine gate is the optimal admission criterion versus, e.g., a residual-norm threshold, or (iii)~the regime in which an SC-realized $W_b$ at $L_{\text{comp}}{<}256$ remains accurate enough to keep the cumulative compensation profitable.
The breakdown we observe at $\bar L{=}96$ (Section~\ref{sec:exp_classification}, ``Open questions'') is consistent with the SC residual exiting the linear-compensation regime, but a formal characterization of where and why is left to future work.

\clearpage
\subsection{Image Generation Samples}
\label{app:image_samples}

\begin{figure*}[h]
\centering
\begin{minipage}{0.49\textwidth}
\centering
\scriptsize Full precision\\[-1pt]
\includegraphics[width=\linewidth]{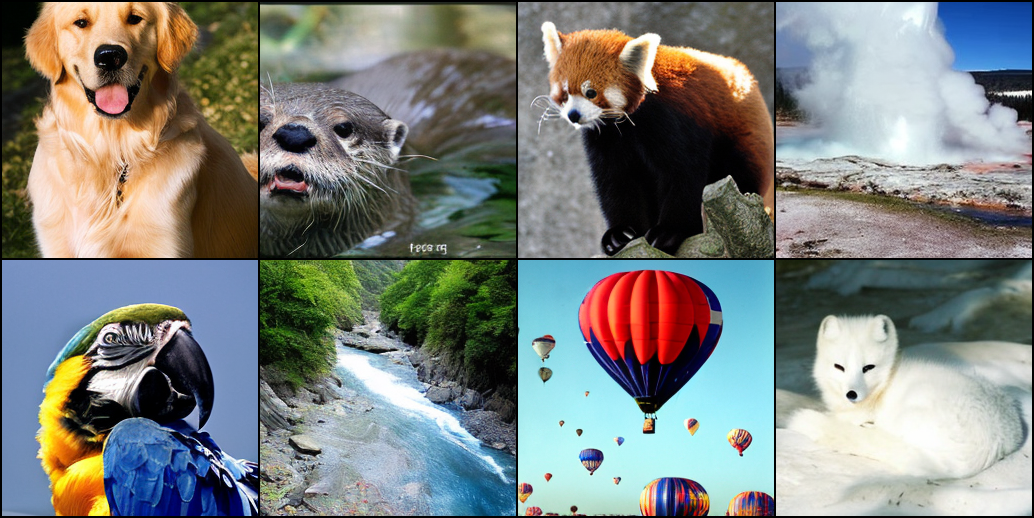}
\end{minipage}\hfill
\begin{minipage}{0.49\textwidth}
\centering
\scriptsize SC, $L=32$\\[-1pt]
\includegraphics[width=\linewidth]{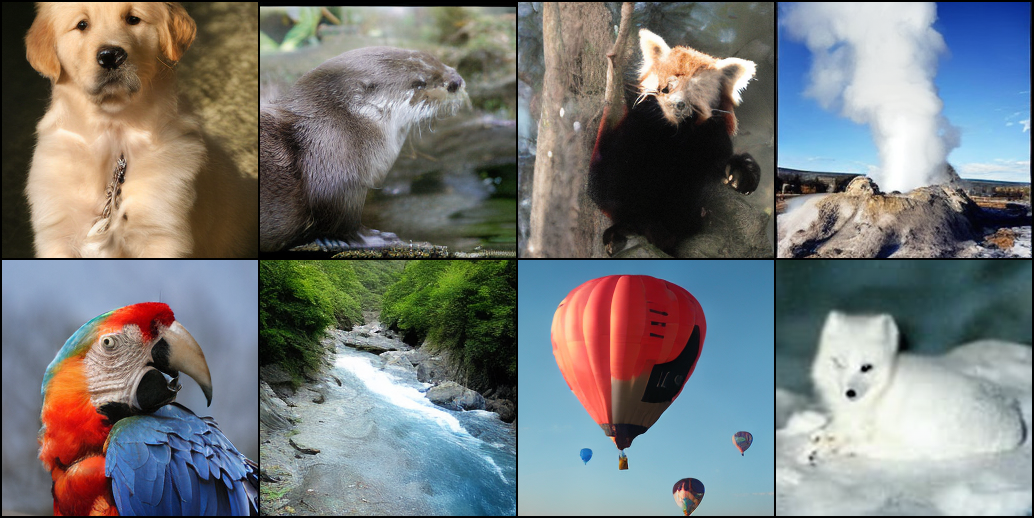}
\end{minipage}

\vspace{3pt}
\begin{minipage}{0.49\textwidth}
\centering
\scriptsize SC, $L=48$\\[-1pt]
\includegraphics[width=\linewidth]{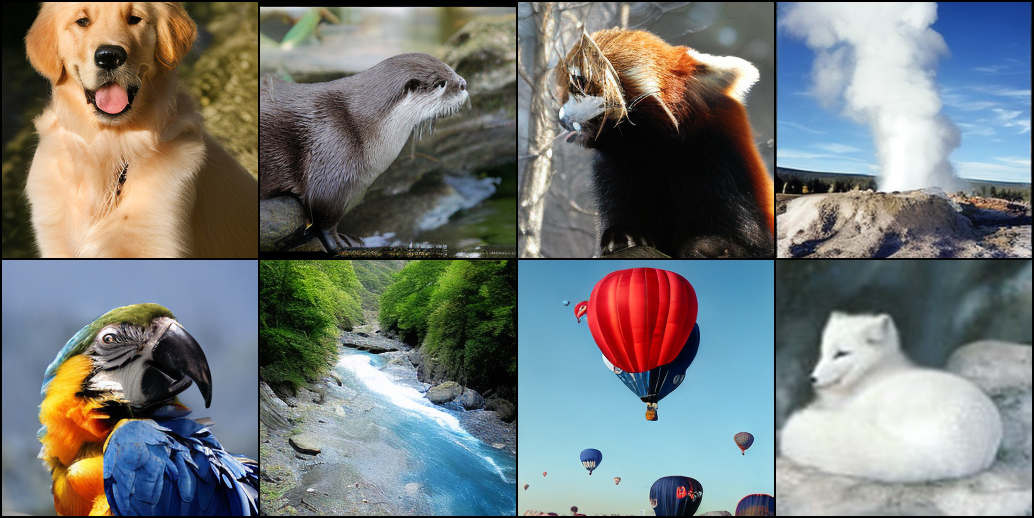}
\end{minipage}\hfill
\begin{minipage}{0.49\textwidth}
\centering
\scriptsize SC, $L=64$\\[-1pt]
\includegraphics[width=\linewidth]{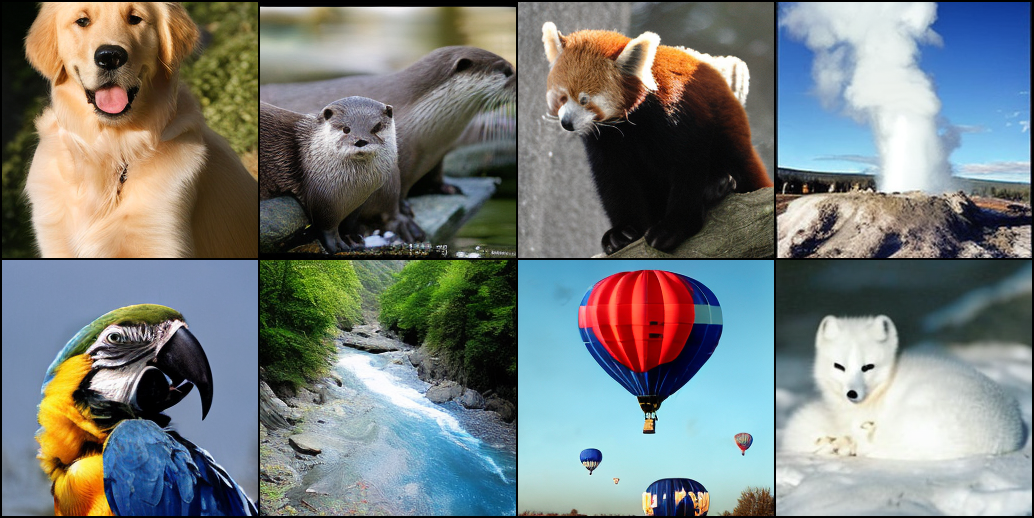}
\end{minipage}

\vspace{3pt}
\begin{minipage}{0.49\textwidth}
\centering
\scriptsize SC, $L=96$\\[-1pt]
\includegraphics[width=\linewidth]{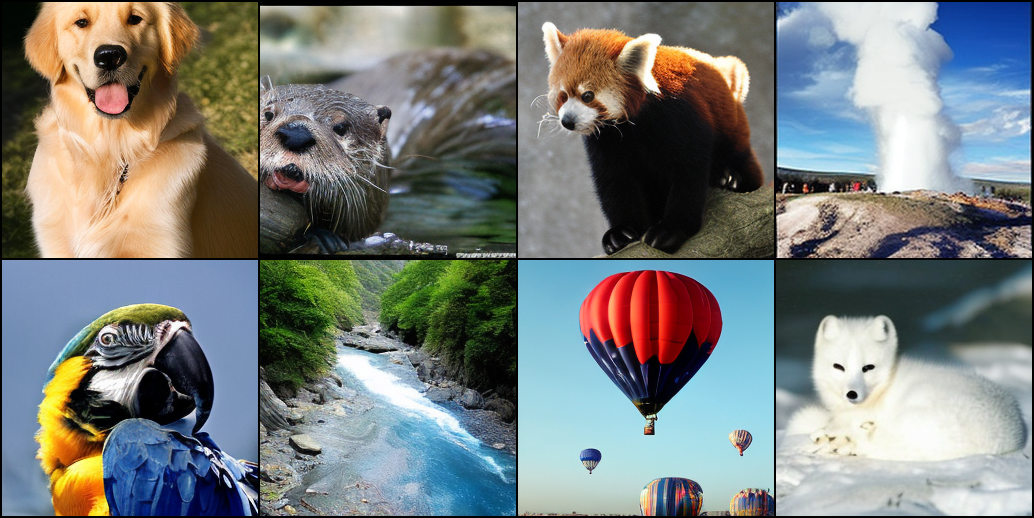}
\end{minipage}\hfill
\begin{minipage}{0.49\textwidth}
\centering
\scriptsize SC, $L=128$\\[-1pt]
\includegraphics[width=\linewidth]{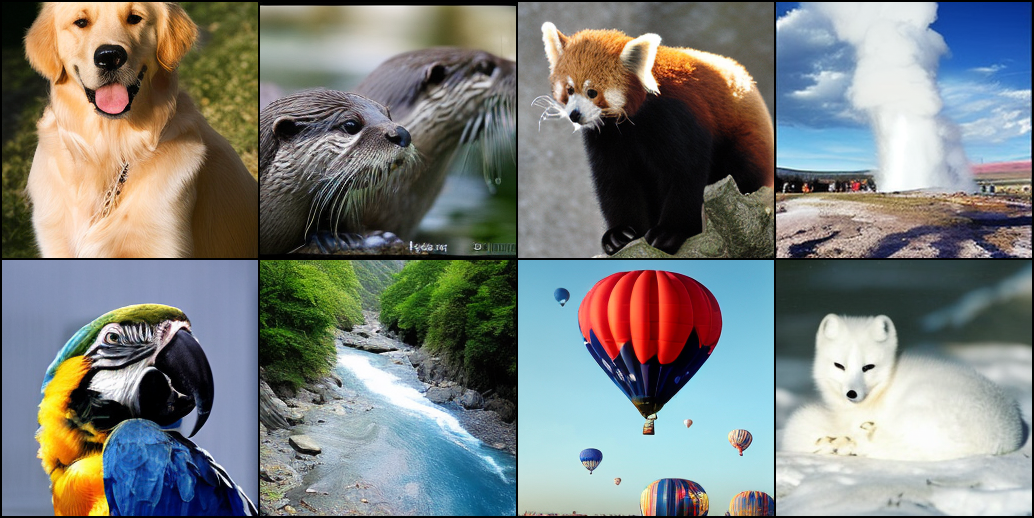}
\end{minipage}
\caption{Sample images generated by full precision and SC with uniform stream lengths $L\in\{32,48,64,96,128\}$. The eight prompts are: ``golden retriever'', ``otter'', ``lesser panda'', ``geyser'', ``macaw'', ``valley'', ``balloon'', and ``arctic fox''.}
\label{fig:image_generation_samples}
\end{figure*}

Figure~\ref{fig:image_generation_samples} shows qualitative samples for the image-generation setting in Section~\ref{sec:exp_image_generation}.

\clearpage

\end{document}